\pdfoutput=1

\documentclass[11pt,a4paper]{article}

\usepackage[preprint]{acl}

\usepackage{times}
\usepackage{latexsym}

\usepackage[T1]{fontenc}

\usepackage[utf8]{inputenc}
\usepackage[german,english]{babel}

\usepackage{microtype}

\usepackage{inconsolata}

\usepackage{graphicx}

\usepackage{booktabs}

\usepackage{listings}
\usepackage{xcolor}

\usepackage{array}

\title{Integration of LLM Quality Assurance into an NLG System}

\author{Ching-Yi Chen \and Johanna Heininger \and Adela Schneider\\
{\bf Christian Eckard} \and {\bf Andreas Madsack} \and {\bf Robert Wei{\ss}graeber} \\
AX Semantics, Stuttgart, Germany \\
{\tt \{firstname.lastname\}@ax-semantics.com}}

\begin{document}
\maketitle
\begin{abstract}
In this paper, we present a system that uses a Large Language Model (LLM) to perform grammar and spelling correction as a component of Quality Assurance (QA) for texts generated by NLG systems, which is important for text production in real-world scenarios. Evaluating the results of the system on work-in-progress sports news texts in three languages, we show that it is able to deliver acceptable corrections. 


\end{abstract}

\section{Introduction}


Improving text quality is essential for the output of natural language generation (NLG) systems. It has been shown that reporting overall performance metrics without identifying and analyzing errors is not a sufficient way to improve NLG systems \cite{vanmiltenburg2021underreporting}.
To bridge the gap in achieving text quality, existing methods such as external QA tools like Grammarly\footnote{\href{www.grammarly.com}{www.grammarly.com}} and frameworks leveraging Large Language Models (LLMs) for text amendments \cite{gou2023critic}, offer valuable assistance but are limited to correcting one text at a time.


We therefore propose an innovative approach to integrate QA in a real-world NLG application. 
The scope of QA in this paper is not only to find spelling and grammatical errors, but also to correct the source of the error in the NLG system, also with the aim of avoiding such or similar errors for subsequent text generations. This is also due to the fact that it is not possible to correct each singular text in a real-life application. With this vision, our LLM QA system is designed to suggest corrections, explain errors, and trace the original errors. Ultimately, users make the final decision to accept or reject suggestions. We perform LLM-based evaluations \cite{fu2023gptscore} to answer the question of whether our LLM QA system effectively improves text quality, which we found it partially does.



\section{Background}


While the discussions on the evaluation of NLG Systems focus on addressing complex issues related to the specification and evaluation of different text quality dimensions \citep{howcroft-etal-2020-twenty, li2022faithfulness}, we consider only the dimension of grammar and spelling correctness for creating this component of a QA system. Our overarching concept that will be developed at a later stage  is to perform quality assurance as an iterative process \citep{du-etal-2022-read} while reviewing each quality dimension in separate process loops. The approach of this first QA component is similar to these types of Grammar Error Correction (GEC) systems that label and explain errors, and build the backbone for editing assistants for individual writing such as Grammarly. Incorrect passages are  marked and supplemented with error types and suggestions for improvement. Whereas automated improvement without intermediate human checking is not the aim.  

While LLMs have already been proven to be effective in other evaluation contexts, they have not yet been sufficiently studied in the field of GEC \citep{loem-etal-2023-exploring}. However, there is increasing evidence that they can produce good results. The biggest challenges are the design of the prompts and the tendency of LLMs to over-correct \citep{wu2023chatgpt,loem-etal-2023-exploring, fang2023chatgpt}.


\section{System Overview}

\begin{figure*}[ht!]
    \centering
    \includegraphics[width=1\textwidth]{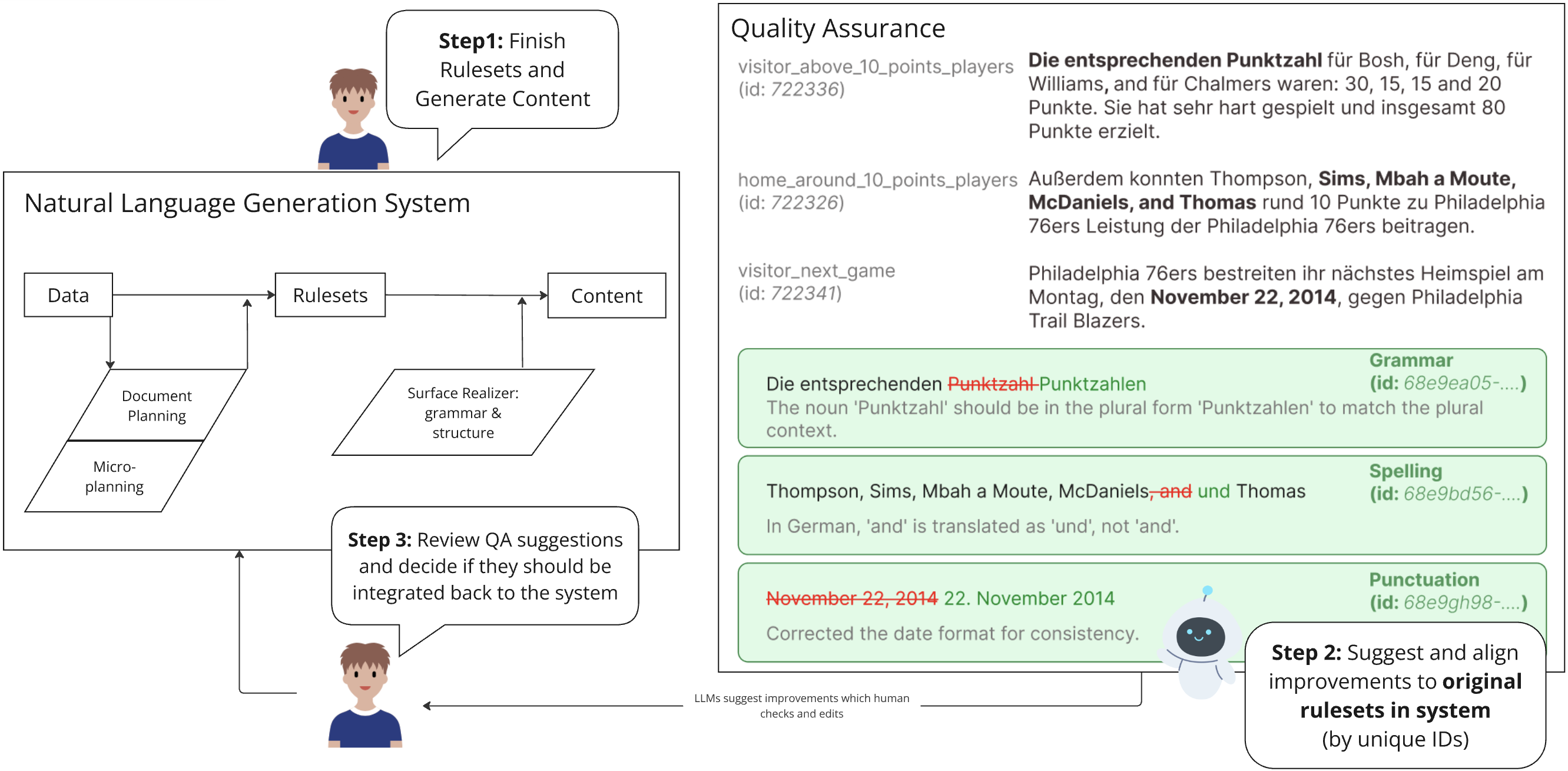}
    \caption{Overview of integrating LLM QA into an NLG system. It demonstrates a human-in-the-loop workflow; the text quality is improved iteratively by repeating steps 2 (LLM suggests corrections) and 3 (human editor checks and carries out corrections)}
    \label{fig:example_image}
    \centering
\end{figure*}

\subsection{Data-to-Text System}
A data-to-text NLG system receives as input (a) data and (b) a specification of how to generate the resulting human language text.
Such a system can be used to automate journalistic texts, financial reports, or product descriptions in e-commerce.
For all of these kinds of texts, assuring the correctness of the system's output is essential.

The NLG generation process has three fundamental steps \citep{reiter/dale:2000}: Document Planning, Micro Planning, and Surface Realization. Our system focuses on Surface Realization, while it enables users to determine planning steps themselves by choosing which and how data fields are used and which sentences are part of their narrative. We call the configurations that are done by users a rule set, which is used for both surface realization and generating text based on the data it is given (see the left side of Figure~\ref{fig:example_image}).

Text coming from data or text having to agree with data is represented in variable units that each have a unique ID and whose grammar can be configured by users (e.g. number, case, or tense). See, for example, these two sports reporting sentences below, which come from two different data inputs:

1) \textbf{Fortuna} won with \textbf{2} \textit{goals} against \textbf{Borussia}.

2) \textbf{Dynamo} won with \textbf{1} \textit{goal} against \textbf{Hertha}. 

The bold words come from data, whereas the italic ones are connected to the data, i.e. the variable unit \textbf{2} is a data point (e.g. number\_of\_goals=2) and the variable unit \textit{goal} is configured to grammatically agree with that number. Therefore \textit{goal} is pluralized in 1), but stays in singular in 2).

\subsection{LLM for Quality Assurance}

In this paper, we show how our existing NLG system \citep{weissgraeber-madsack-2017-working} can be optimized by utilizing the expertise of LLMs as domain experts to assess generated texts based on the quality criteria of correctness, i.e. grammar and spelling. 

We built a framework that instructs an LLM to identify potential errors in a representative subset of the rendered text of our NLG system generated from basketball data and give suggestions on how to fix them. For example, we prompt the LLM to correct ten basketball match result texts. The suggestions are linked to the original rule set via unique IDs to be able to point to the exact spot where an error occurred.

Users can then check the suggestions and decide whether to re-integrate the corrections into the system (see right side of Figure~\ref{fig:example_image}).
These corrections potentially improve not only one text but many more texts that are generated subsequently.

In this human-in-the-loop scenario, we assume that the QA performed by the LLM helps users to make effective decisions to fix errors and ultimately receive better texts from the optimized system.

A shortened version of our original \textbf{prompt} reduced to its essential aspects would look like this:

\vspace{5px}
\noindent 
\fbox{\parbox{\linewidth}{
Identify all errors in the given text.

Provide suggestions to enhance correctness.

Consider grammar, punctuation, and spelling. 

Map the erroneous sections to their IDs.
}}

\section{Experiment}
We conducted experiments to answer the research question: \textbf{Does our LLM QA system effectively improve text quality?}
We answer the question by evaluating 1) the system's ability to identify ungrammatical text and 2) the system's ability to provide high-quality suggestions and authentic text revisions.
   
\subsection{Experimental Setups}
Our NLG system is capable of machine translation. To come up with a test dataset, we have built English rule sets and performed machine translation for English\textrightarrow French, English\textrightarrow German, and English\textrightarrow Spanish. Automatically translating rule sets and text will inevitably introduce errors, including unlocalized punctuation, mismatched grammatical agreements, etc. We fix these errors by having the LLM QA system identify errors and suggest revisions for each rule set.

\subsubsection{Data Preparation}
We use the Sport-Sett: Basketball dataset\footnote{\href{https://huggingface.co/datasets/GEM/sportsett_basketball}{https://huggingface.co/datasets/GEM/sportsett\_basketball}} \cite{thomson-etal-2020-sportsett} to generate basketball match reports using our NLG system. Once translated into German, French, and Spanish, each language will have 10 generated reports. Table~\ref{table: recall_and_precision} shows the total number of words in all generated reports. We then collect their rule sets as input for the LLM QA system. Eventually, we run evaluations for the outputs of the LLM QA system. 

\subsubsection{Design of LLM Evaluations}

\lstset{
  basicstyle=\ttfamily\footnotesize,
  columns=fullflexible,
  frame=single,
  breaklines=true,
  postbreak=\mbox{\textcolor{red}{$\hookrightarrow$}\space},
  showstringspaces=false,
  linewidth=1.0\linewidth,
  extendedchars=true,
  inputencoding=utf8,
  literate=%
  {ü}{{\"u}}1
}

\begin{figure*}[ht!]
    \centering
    \begin{lstlisting}
    {   "id": 722336,
        "revised_text": "Die entsprechenden Punktzahlen für Afflalo, für ...",
        "containers": [{
                "id": "68e9ea05-8b5a-4252-b267-ef9d5944e3cc",
                "revised_text": "Die entsprechenden Punktzahlen",
                "explanation": "'Punktzahl' should be plural 'Punktzahlen' to match the plural...",
                "intention": "grammar",
                "text": "Die entsprechenden Punktzahl"
            }, ...],
        "text": "Die entsprechenden Punktzahl für Afflalo..." }
    \end{lstlisting}
    \caption{The partial JSON output of our LLM QA system. The id, text, and containers came from our rule sets in the NLG system. The revised text, explanation, and intention came from the LLM QA system. The revised text, explanation, and text fields are used in the LLM-based evaluation.}
    \label{fig:llm_qa_system_output}
\end{figure*}

The LLM QA system's results were evaluated using another LLM, referred to as the \textbf{LLM Evaluator}. We assess performances of the LLM QA system via two sets of metrics.

1) \textbf{Precision and Recall}: evaluate the system's ability to identify ungrammatical text. 

2) \textbf{Suggestion Quality and Improvement Proportion}: evaluate the system's ability to provide high-quality suggestions and authentic text revisions.

The LLM Evaluator will act as the ground truth for calculating precision and recall metrics. It also assesses the suggestion quality and revision authenticity provided by the LLM QA system in Figure \ref{fig:llm_qa_system_output}. All LLM-based tasks are carried out by GPT-4 (gpt-4-0125-preview) and via its API\footnote{\href{https://openai.com/index/new-embedding-models-and-api-updates/}{https://openai.com/index/new-embedding-models-and-api-updates/}} to prevent data contamination \cite{balloccu-etal-2024-leak}. 

For the LLM QA system, we set the temperature to 0.5, and for the LLM Evaluator to 0 to make it more deterministic. We run all evaluations ten times for each language to calculate mean scores, which provides a more reliable measure by reducing variability. 

\subsection{Result Analysis}

\subsubsection{Precision and Recall}

\begin{table}[ht!]
    \centering
    \small
    \begin{tabular}{lrrrr}
        \toprule
        Language & Words &  Ungram. & Precision & Recall \\
        \midrule
        French & 4579 & 86\% & 94\% & 92\% \\
        German & 3944 & 47\% & 96\% & 55\% \\
        Spanish & 4449 & 65\% & 98\% & 85\% \\
        \bottomrule
    \end{tabular}
    
    \caption{The total number of  words in all generated texts, the proportion of ungrammatical sentences in the texts (identified by the LLM Evaluator), and the precision and recall of the ungrammatical text.}
    \label{table: recall_and_precision}
\end{table}

To evaluate whether the system could identify all ungrammatical text parts, we first calculate the proportion of ungrammatical sentences, because this provides context for interpreting precision and recall. Precision measures the \textbf{relevance}
of the system in identifying ungrammatical text. Recall measures the \textbf{completeness}
of the system in identifying ungrammatical text.

In Table~\ref{table: recall_and_precision}, the relatively high rate of identified ungrammatical text for French (86\%) indicates a significant presence of actual ungrammatical text. Our LLM QA system identifies these errors effectively, as shown by high precision and recall. However, low recall (55\%) for German suggests that the system misses 45\% of ungrammatical text.

\subsubsection{Suggestion Quality and Improvement Proportion}

\begin{table}[ht!]
    \small
    \centering
    \begin{tabular}{lrr}
        \toprule
        Language & Suggestion Qual. & Improvement Pro. \\
        \midrule
        French & 89.19 & 59\% \\
        German & 90.38 & 31\% \\
        Spanish & 94.67 & 62\% \\
        \bottomrule
    \end{tabular}
    \caption{The LLM Evaluator rates the score for Suggestion Quality and checks the authenticity of the revised text.}
    \label{table: quality_of_suggestion_and_reivsion}
\end{table}

The Suggestion Quality in Table~\ref{table: quality_of_suggestion_and_reivsion} is an average score of the quality of suggestions provided by the LLM QA system (from 0 to 100, 100 being best).
Improvement Proportion indicates the proportion of the revised text
that the LLM Evaluator identifies as an improvement over the original text. Despite the high Suggestion Quality for the three languages, the low Improvement Proportion score implies that revisions suggested by the LLM QA system are not consistently perceived as authentic improvements, particularly evident in German. 

\subsection{Discussion}
In general, the LLM QA system demonstrates high precision and favorable ratings for the quality of suggestions. The significant challenges remain in \textbf{the authenticity of text revision}, which is reflected in the poor "Improvement Proportion" in Table~\ref{table: quality_of_suggestion_and_reivsion}. We observe the variation of evaluation results across three languages. Of particular concern is German, with a recall of 0.55 and an Improvement Proportion of 31, indicating its shortcomings in both error detection and quality of revision. When reviewing our system's suggestions, we notice that the majority of hallucinations are caused by our LLM QA system suggesting an unwanted revision to the error-free text or inaccurate revision (as shown in Figure \ref{fig:output} in Appendix). Therefore, we propose a human-in-the-loop workflow, allowing users to accept or reject suggestions.

Potential optimizations could include implementing a dynamic prompt tailored to the language and type of text. Using the same prompt for all languages can cause the oversight of many ungrammatical sentences and low recall scores. Furthermore, we could introduce few-shot prompting to provide clear examples of what constitutes a high-quality revision to optimize Improvement Proportion.

\section{Limitations and Potential Risks}

Due to time constraints, we are \textbf{not evaluating the full human-in-the-loop system} which consists of iterative feedback, and automated correction of grammatical and spelling errors as a stopping mechanism. Instead, the focus of our evaluation is on the quality and precision of a single correction task.

In terms of scope, our \textbf{evaluation} is only \textbf{based on a single dataset} (basketball match reports) with 10 example texts in three languages (German, Spanish, French), that were generated via translation. They might therefore have different types and distributions of errors than texts that are written from scratch.

We \textbf{assume that an LLM (GPT-4) can evaluate itself} in terms of grammatical correctness and spelling (see for example \citep{lin-chen-2023-llm}). However, human evaluation is needed to improve the evaluation completeness.

Additionally, LLM-QA \textbf{inherits the limitations that are present in GPT-4}, for example, 'hallucination', and varying performance for some languages \citep{openai2024gpt4}, e.g. Finnish and Hungarian (however, to our knowledge, no official list of supported languages exists). This ultimately contributes to the underexposure of those specific languages.

\section{Conclusion}

We contribute a system for Quality Assurance of NLG systems based on LLMs. Our main focus lies on how to prompt the LLM, as well as input and output processing to embed the QA system into the logic of an NLG system. We further show how to build an LLM-based evaluation for the QA system. First results are promising, but also highlight areas for improvement, for example to tailor the prompt for better precision and recall.

As a future improvement, we want to offer a way for users to accept suggestions from the LLM and semi-automatically integrate the corrections back into the system's configuration. This means either showing a link to the place that should be corrected or offering a button that automatically applies the corrections.
Furthermore, we plan to additionally include other text quality features like engagement, clarity, and delivery. Checking for these aspects of readability should help users to get even more inspiration for improving their texts.  

\newpage


\bibliography{custom}

\newpage
\onecolumn
\appendix

\section{Appendix}

\label{sec:appendix}
\subsection{Example Prompts}

Using German as an example, we present the following prompts: the prompt used for the LLM QA system in Figure~\ref{fig:prompt-for-LLM-QA-system}, the ground truth generation prompt for calculating Precision and Recall in Figure~\ref{fig:prompt-ground-truth}, the prompt to rate Suggestion Quality in Figure~\ref{fig:prompt-suggestion-quality}, and the prompt to examine the authenticity of revised text in Figure~\ref{fig:prompt-reivsion-quality}.

\lstset{
  basicstyle=\ttfamily\small,
  columns=fullflexible,
  frame=single,
  breaklines=true,
  postbreak=\mbox{\textcolor{red}{$\hookrightarrow$}\space},
  showstringspaces=false,
  linewidth=1.0\linewidth,
  extendedchars=true,
  inputencoding=utf8,
  literate=%
  {Ö}{{\"O}}1
  {Ä}{{\"A}}1
  {Ü}{{\"U}}1
  {ß}{{\ss}}1
  {ü}{{\"u}}1
  {ä}{{\"a}}1
  {ö}{{\"o}}1
}
\begin{figure*}[ht!]
\centering
\begin{lstlisting}
system_msg = "Identify all problematic texts, and then provide suggestions to enhance 
the correctness of given GERMAN texts. You are designed to output JSON."

user_msg= "
While analyzing, consider the following aspects of correctness:

    - Grammar: Ensure the text uses correct syntax, number, and tenses in GERMAN, 
    with attention to gender agreement. Pay attention to case inflection 
    for accompanying determiners, adjectives, numerals, and pronouns.
    - Punctuation: Check for the proper use of punctuation to structure sentences.
    - Spelling: Verify the correct spelling of words.


Please adhere strictly to the original name entities, URLs, markup (e.g., <b>, #, *, |), 
and special characters like newlines or umlauts. Refrain from suggesting any unnecessary alterations.
        Here is the text you need to analyze:

        \"\"\"{all_plain_text}\"\"\"

        After your analysis, map the sections that need improvement to their IDs 
        in the provided ruleset JSON:

        \"\"\"{rule_set}\"\"\""
\end{lstlisting}
\caption{The prompt for the LLM QA system.}
\label{fig:prompt-for-LLM-QA-system}
\end{figure*}

\begin{figure*}[ht]
    \centering
    \begin{lstlisting}
    system_msg = "Answer if given text is correct in terms of grammar and spelling in {language}. 
    The answer needs to be True or False as string. Return only the answer."
    
    user_msg = "Here are the text '''{text}'''." 
    \end{lstlisting}
    \caption{The prompt for generating ground truth in order to calculate precision and recall.}
    \label{fig:prompt-ground-truth}
\end{figure*}

\begin{figure*}[ht]
    \centering
    \begin{lstlisting}
    system_msg = "Rate the following revised text and explanation, given a text in terms of grammar 
    and spelling in GERMAN. The rating needs to be the number between 0 and 100. 
    Return only the rating."
    
    user_msg = "Here are the original text '''{text}''', explanation '''{explanation}''', 
    and the revised text '''{revised_text}'''." 
    \end{lstlisting}
    \caption{The prompt to rate suggestion quality.}
    \label{fig:prompt-suggestion-quality}
\end{figure*}

\begin{figure*}[ht]
    \centering
    \begin{lstlisting}
    system_msg = "Decide if the revised text is better than the original text 
    in terms of spelling and grammar. Return: [true, false]"
    
    user_msg = "original text: '''{text}''', revised text: '''{revised_text}'''." 
    \end{lstlisting}
    \caption{The prompt to validate revision authenticity.}
    \label{fig:prompt-reivsion-quality}
\end{figure*}

\section{Data Formats}
Figure~\ref{fig:debug-output} is an example of the input data for the LLM QA system, which is the rule set of our NLG system. Figure~\ref{fig:output} shows the output of our LLM QA system. These two examples are aligned with each other, where the suggestions in Figure~\ref{fig:output} are specific to fix the errors in Figure~\ref{fig:debug-output}.

\begin{figure*}[ht!]
    \centering
    \begin{lstlisting}
    {
        "id": 722329,
        "name": "home_double_double",
        "text": "Außerdem zeigte Wroten mit seinem Doubl-Double-Leistung 
        seine Multi-Talente Allround-Fähigkeit.",
        "containers": [
            ...
            {
                "text": "zeigte",
                "id": "3964ee99-5f41-49c1-8956-231ae6c9487e"
            },
            {
                "text": " "
            },
            {
                "text": "Wroten",
                "id": "6ff6b9a9-0fab-4612-ae46-f81cf281a158"
            },
            {
                "text": " "
            },
            {
                "text": "mit seinem Double-Double-Leistung",
                "id": "0202ccfc-f604-4471-aa17-7f8ec5fce02b"
            },
            {
                "text": " "
            },
            {
                "text": "seine Multi-Talente Allround-Fähigkeit",
                "id": "c0c6a367-eb58-49ec-b7ce-cd4085711b5b"
            },
            ...
        ]
    }
    \end{lstlisting}
    \caption{The data format of the partial rule set which is also used as the input of the LLM QA system.}
    \label{fig:debug-output}
\end{figure*}

\begin{figure*}[ht]
    \centering
    \begin{lstlisting}
    {
        "id": 722329,
        "name": "home_double_double",
        "revised_text": "Außerdem zeigte Wroten mit seiner Double-Double-Leistung 
        seine Multi-Talente und Allround-Fähigkeiten.",
        "containers": [
            {
                "id": "0202ccfc-f604-4471-aa17-7f8ec5fce02b",
                "revised_text": "mit seiner Double-Double-Leistung",
                "explanation": "The possessive pronoun 'seiner' must agree in gender (masculine) 
                and case (genitive) with the noun 'Leistung'.",
                "intention": "grammar",
                "text": "mit seinem Double-Double-Leistung"
            },
            {
                "id": "c0c6a367-eb58-49ec-b7ce-cd4085711b5b",
                "revised_text": "seine Multi-Talente und Allround-Fähigkeiten",
                "explanation": "The noun 'Fähigkeit' should be in the plural form to match 
                'Multi-Talente', and the connection of the phrases should be made clear with 'und'.",
                "intention": "grammar",
                "text": "seine Multi-Talente Allround-Fähigkeit"
            }
        ],
        "text": "Außerdem zeigte Wroten mit seinem Double-Double-Leistung seine
        Multi-Talente Allround-Fähigkeit."
    }
    
    \end{lstlisting}
    \caption{LLM QA system output for German (with unique IDs), aligning with the input in Figure~\ref{fig:debug-output}. The first explanation in the containers shows how the LLM QA system hallucinates. The reasonable explanation could be "\textit{The possessive pronoun 'seiner' must agree in gender (feminine) and case (dative) with the noun 'Leistung'.}".  }
    \label{fig:output}
\end{figure*}

\section{Quality Assurance: Original vs. Revised Text}
Table~\ref{fig:before and after revision} compares the original text to the revised text and the corresponding explanations. The LLM QA system generates the revised text and corresponding explanations. The first row aligns with the output example of our LLM QA system shown in Figure~\ref{fig:output}.

\begin{table*}[ht!]
    \centering
    \begin{tabular}{| p{3.5cm} | p{3.5cm} | p{7cm} |}
    \hline
    \textbf{Original Text} & \textbf{Revised Text} & \textbf{Explanations} \\ \hline
    Außerdem zeigte Wroten mit seinem Double-Double-Leistung seine Multi-Talente Allround-Fähigkeit. & Außerdem zeigte Wroten mit seiner Double-Double-Leistung seine Multi-Talente und Allround-Fähigkeiten. & 1. The possessive pronoun 'seiner' must agree in gender (masculine) and case (genitive) with the noun 'Leistung'. \newline 2. The noun 'Fähigkeit' should be in the plural form to match 'Multi-Talente', and the connection of the phrases should be made clear with 'und'. \\ \hline
    Während des Spiels haben Wroten and Davies mehr als 10 Punkte erzielt und diese 2 bemerkenswerten Spieler hat insgesamt 39 Punkte gemacht. & Während des Spiels haben Wroten und Davies mehr als 10 Punkte erzielt und diese zwei bemerkenswerten Spieler haben insgesamt 39 Punkte gemacht. & 1. In German, 'and' is translated as 'und', not 'and'. \newline 2. The numeral '2' should be written as 'zwei' in full words in German texts. \newline 3. The verb 'haben' should agree in number with the plural subject 'Spieler'. \\ \hline
    Außerdem konnten Thompson, Sims, Mbah a Moute, McDaniels, and Thomas rund 10 Punkte zu Philadelphia 76ers Leistung der Philadelphia 76ers beitragen. & Außerdem konnten Thompson, Sims, Mbah a Moute, McDaniels und Thomas rund 10 Punkte zur Leistung der Philadelphia 76ers beitragen. & 1. In German, 'and' is translated as 'und', not 'and'. \newline 2. The preposition 'zu' should be combined with the definite article 'der' to form 'zur', to match the dative case required by 'beitragen'. \\ \hline
    Die entsprechenden Punktzahl für Bosh, für Deng, für Williams, and für Chalmers waren: 30, 15, 15 and 20 Punkte. Sie hat sehr hart gespielt und insgesamt 80 Punkte erzielt. & Die entsprechenden Punktzahlen für Bosh, Deng, Williams und Chalmers waren: 30, 15, 15 und 20 Punkte. Sie haben sehr hart gespielt und insgesamt 80 Punkte erzielt. & 1. The noun 'Punktzahl' should be in the plural form 'Punktzahlen' to match the plural context. \newline 2. In German, 'and' is translated as 'und', not 'and'. \newline 3. The verb form 'haben' should agree with the plural subject 'Sie' to correctly form the past tense. \\ \hline
    Außerdem insgesamt 12 Dreipunktspiele erfolgreich abgeschlossen Außerdem hat Bosh, Deng, Williams, Cole, Chalmers, Napier, and Ennis während des Spiels . & Außerdem wurden insgesamt 12 Dreipunktspiele erfolgreich abgeschlossen. Bosh, Deng, Williams, Cole, Chalmers, Napier und Ennis haben während des Spiels... & 1. The sentence is missing a verb to complete the action. 'Wurden' should be added at the beginning to correctly form the passive voice. \newline 2. In German, 'and' is translated as 'und', not 'and'. The sentence also seems incomplete and might require further context to be fully corrected. \\ \hline
    \end{tabular}
     \caption{The original text versus the revised text (generated by the LLM QA system) along with corresponding explanations (also generated by the LLM QA system). The first row aligns with the output example of our LLM QA system shown in Figure \ref{fig:output}.}
    \label{fig:before and after revision}
\end{table*}

\section{Evaluation}

Table~\ref{table:76ers_heat_game} presents the partial sanity check, which assesses the alignment between the ground truth (from LLM Evaluator) and human review. It demonstrates how closely the decisions of the LLM Evaluator correspond to those of the human evaluator. Table~\ref{table:histograms} displays the histogram of evaluations, indicating that a regression to the mean occurs and that variations are minimal.

\begin{table*}[ht]
    \centering
    \begin{tabular}{| l | l | p{10cm} |}
    \hline
    \textbf{Human} & \textbf{LLM Evaluator} & \textbf{Sentence} \\
    \hline
    Correct & Correct & \#\#\# Philadelphia 76ers vs. Miami Heat \\
    \hline
    \textcolor{red}{Incorrect} & \textcolor{red}{Correct} & Das Spiel zwischen den Philadelphia 76ers und \textcolor{orange}{den Miami Heat} fand am Samstag (01. November 2014) im ausverkauften Wells Fargo Center (Pennsylvania) statt. \\
    \hline
    Correct & Correct & Rund 20000 begeisterte Fans kamen am 1. Spieltag der Saison 2014 nach Philadelphia und füllten das Stadion komplett. \\
    \hline
    Correct & Correct & Die Gastmannschaft aus Miami gewann gegen die Heimmannschaft mit 114:96. \\
    \hline
    \textcolor{red}{Incorrect} & \textcolor{red}{Correct} & Der beste Spieler des Spiels war zweifelsohne Chris Bosh von \textcolor{orange}{den Miami Heat}. Er führte das Team mit 30 Punkten, 8 beeindruckenden Rückprallern, 4 präzisen Vorlagen und 2 spielverändernden Steals zum Sieg. \\
    \hline
    Correct & Correct & Im Gegenteil, die Philadelphia 76ers konnten trotz der beeindruckenden Leistungen ihres Spitzenspielers keinen Sieg erringen. Tony Wroten war mit 21 Punkten der Topscorer des Teams. \\
    \hline
    Correct & Correct & \#\#\# Philadelphia 76ers Höhepunkte \\ \hline
    Incorrect & Incorrect & Tony Wroten unternahm im gesamten Spiel 11 Feldversuche und traf \textcolor{blue}{6 ihnen}, was einer Erfolgsquote von 55 \% entspricht. \\ \hline
    Correct & Correct & Er erwies sich auch als Freiwurfschütze, denn er traf 8 seiner 11 Versuche und erzielte damit eine Freiwurfquote von 73 \%. \\ \hline
    Incorrect & Incorrect & Außerdem zeigte Wroten mit \textcolor{blue}{seinem} Double-Double-Leistung seine Multi-Talente Allround-Fähigkeit. \\ \hline
    Incorrect & Incorrect & Während des Spiels haben Wroten \textcolor{blue}{and} Davies mehr als 10 Punkte erzielt und diese \textcolor{green}{2} bemerkenswerten Spieler \textcolor{blue}{hat} insgesamt 39 Punkte gemacht. \\ \hline
    Incorrect & Incorrect & Außerdem konnten Thompson, Sims, Mbah a Moute, McDaniels, \textcolor{blue}{and} Thomas rund 10 Punkte \textcolor{blue}{zu Philadelphia 76ers Leistung der Philadelphia 76ers} beitragen. \\ \hline
    \textcolor{red}{Correct} & \textcolor{red}{Incorrect} & \#\#\#\# Miami Heat Highlights \\ \hline
    Correct & Correct & Chris Bosh hat in diesem Spiel 17 Feldtore erzielt, von denen er 9 erfolgreich traf, was einer Trefferquote von 53 \% entspricht. \\ \hline
    Correct & Correct & Auch von der Freiwurflinie war er effektiv: Er traf 10 seiner 11 Versuche und erzielte damit eine Freiwurfquote von 91 \%. \\ 
    \hline
    \end{tabular}
    \caption{The punctual sanity check assesses the alignment between the LLM Evaluator's ground truth and a human review. The table shows, whether the text was categorized as correct or incorrect and demonstrates that the LLM Evaluator's decision largely corresponds with the judgments of a human (native speaker) evaluator: only 3 out of 15 sentences are labeled differently (highlighted with red). The errors marked by both human and LLM are highlighted with blue, whereas the errors only marked by the LLM are highlighted with green and the ones only marked by the human annotator are orange. Two of the differently rated sentences are caused by the same construction, which is a rather small error; the other one is probably due to the special characters "\#\#\#\#", which the LLM labels as incorrect.}
    \label{table:76ers_heat_game}
\end{table*}

\begin{table*}
    \centering
    \begin{tabular}{|m{3cm}|c|c|c|}
        \hline
        \textbf{Metric} &  \textbf{French}  & \textbf{German} & \textbf{Spanish} \\
        \hline
        Accuracy & \includegraphics[width=0.21\linewidth]{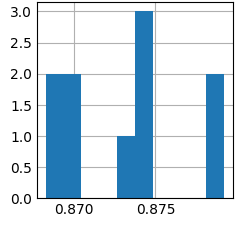} & \includegraphics[width=0.21\linewidth]{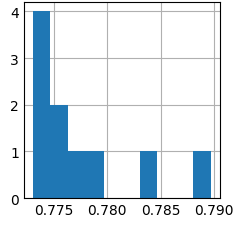} & \includegraphics[width=0.21\linewidth]{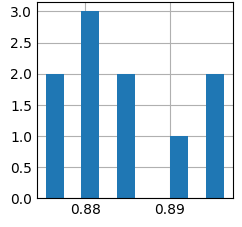}\\
        \hline
        Precision & \includegraphics[width=0.21\linewidth]{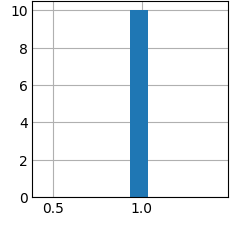} & \includegraphics[width=0.21\linewidth]{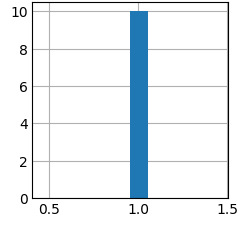} & \includegraphics[width=0.21\linewidth]{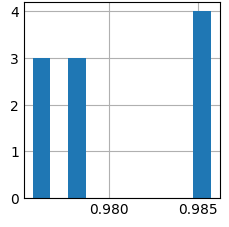}\\
        \hline
        Recall & \includegraphics[width=0.21\linewidth]{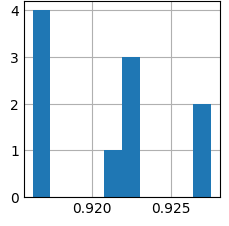} & \includegraphics[width=0.21\linewidth]{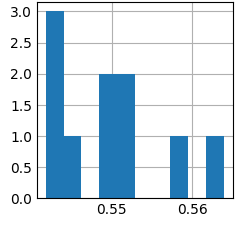} & \includegraphics[width=0.21\linewidth]{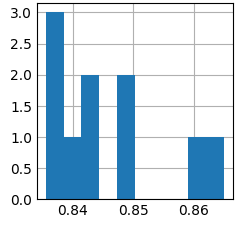}\\
        \hline
        Rate of Ungrammatical Text & \includegraphics[width=0.21\linewidth]{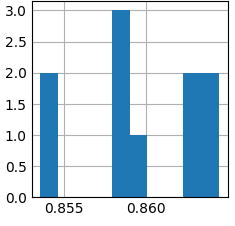} & \includegraphics[width=0.21\linewidth]{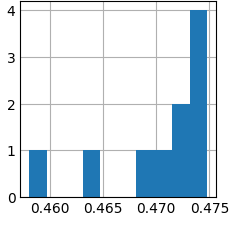} & \includegraphics[width=0.21\linewidth]{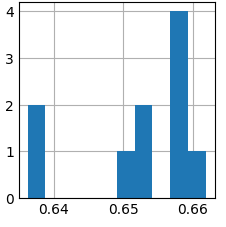}\\
        \hline
        Suggestion Quality & \includegraphics[width=0.21\linewidth]{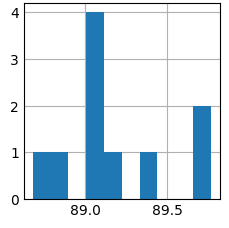} & \includegraphics[width=0.21\linewidth]{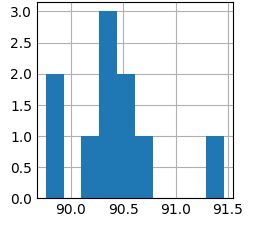} & \includegraphics[width=0.21\linewidth]{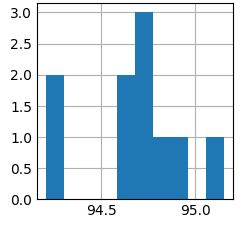}\\
        \hline
        Improvement Proportion & \includegraphics[width=0.21\linewidth]{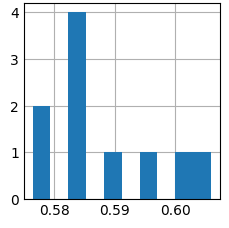} & \includegraphics[width=0.21\linewidth]{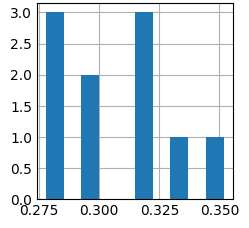} & \includegraphics[width=0.21\linewidth]{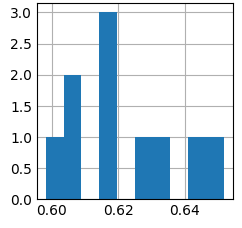}\\
        \hline
    \end{tabular}
    \caption{Histograms of 10 evaluation repetitions. Y-axes are counts, and x-axes are the metric's domain.}
    \label{table:histograms}
\end{table*}

\end{document}